\documentclass{article}

\usepackage[preprint]{neurips_2026}

\usepackage[utf8]{inputenc} 
\usepackage[T1]{fontenc}    
\usepackage{hyperref}       
\usepackage{url}            
\usepackage{booktabs}       
\usepackage{amsfonts}       
\usepackage{nicefrac}       
\usepackage{microtype}      
\usepackage{xcolor}         

\usepackage[pdftex]{graphicx}

\usepackage{mathtools}
\usepackage{amsmath}
\usepackage{amssymb}
\usepackage{pbox}
\usepackage{tabularx}
\usepackage{array}
\usepackage{subcaption}
\usepackage{CJKutf8} 
\usepackage[toc,page,header]{appendix}
\usepackage{minitoc}

\usepackage{xcolor}

\newcommand{\DefinedAs}{\triangleq}
\newcommand{\Reals}{\mathbb R}

\newcommand{\Time}{t}
\newcommand{\Length}{T}

\newcommand{\WordIndex}{u}
\newcommand{\NumWords}{U}

\newcommand{\StepIndex}{i}

\newcommand{\Vocabulary}{\mathcal V}
\newcommand{\VocabularyWithWait}{\bar{\mathcal V}}

\newcommand{\Wait}{W}
\newcommand{\Emit}{E}

\newcommand{\Audio}{\mathbf A}
\newcommand{\AudioVector}{\mathbf a}
\newcommand{\EncodedAudio}{\mathbf E}
\newcommand{\EncodedAudioVector}{\mathbf e}

\newcommand{\DecodedSize}{D}

\newcommand{\WaitEmitSequence}{\Theta}
\newcommand{\WaitEmitStep}{\theta}

\newcommand{\Words}{\mathbf w}
\newcommand{\Word}{w}
\newcommand{\LLMInput}{\mathbf X}
\newcommand{\Step}{y}
\newcommand{\Steps}{\mathbf y}
\newcommand{\StartToken}{\text{\emph{start}}}
\newcommand{\EOS}{\text{$<$EOS$>$}}

\newcommand{\Embedding}{\mathbf z}

\newcommand{\WindowSize}{f}

\newcommand{\StepsFor}{\mathcal Y}
\newcommand{\CompleteStepsFor}{\mathcal Y^+}

\newcommand{\WordOne}{\text{this}}
\newcommand{\WordTwo}{\text{is}}
\newcommand{\WordThree}{\text{quick}}

\newcommand{\Encoder}{e}
\newcommand{\Decoder}{d}
\newcommand{\NonWaitDecoder}{d'}
\newcommand{\Joiner}{j}
\newcommand{\NonWaitJoiner}{j'}
\newcommand{\WaitPolicy}{\pi}

\newcommand{\WaitPenalty}{\kappa}
\newcommand{\WaitPolicyWaitPenalty}{\nu}

\newcommand{\StartDelay}{k}

\newcommand{\DecoderState}{\mathbf{d}}

\newcolumntype{Y}{>{\raggedright\arraybackslash}X}

\title{Streaming Speech-to-Text Translation with~a~SpeechLLM}

\author{
  Titouan Parcollet$^*$ \quad  Shucong Zhang$^*$ \quad Xianrui Zheng$^*$\quad Rogier C.\ van Dalen\thanks{These authors contributed equally. Work performed while Xianrui was an intern at Samsung, AI Center -- Cambridge.}\\\\
  Samsung, AI Center -- Cambridge, United Kingdom  \\
}

\begin{document}

\doparttoc
\faketableofcontents

\maketitle

\begin{abstract}
Normally, a system that translates speech into text consists of separate modules for speech recognition and text-to-text translation.
Combining those tasks into a SpeechLLM promises to exploit paralinguistic information in the speech and to reduce cascaded errors.
But existing SpeechLLM systems are slow since they do not work in a real streaming fashion: they wait for a complete utterance of audio before outputting a translation, or output tokens at fixed intervals, which is not suitable for real applications.
This work proposes an LLM-based architecture for real streaming speech-to-text translation.
The LLM learns not just to emit output tokens, but also to decide whether it has seen enough audio to do so.
The system is trained using automatic alignments of the input speech and the output text.
In experiments on different language pairs, the system achieves a translation quality close to the non-streaming baseline, but with a latency of only 1--2 seconds.

\end{abstract}

\section{Introduction}
\label{section:introduction}

The architecture of systems that translate speech in one language into text in another language is often a cascade of a speech recogniser and a text-to-text translation system.
However, in recent years it has become possible to connect a speech encoder to a large language model, producing a model often called a ``SpeechLLM''.
This removes the explicit representation of the source text, and with it the need to run a search algorithm to infer the source text.
The more important promise of a SpeechLLM is it can use information that would be lost in text, such as prosody, hesitations, and other paralinguistic features \citep{tsiamas-2024-speech}.

However, no SpeechLLM system for speech-to-text translation in the literature does real streaming, where the translation is output as sufficient audio is available to do so.
The only models that output translation tokens as the audio comes in, Bestow \citep{chen-2024-bestow} and FASST \citep{ouyang-2024-fasst}, use a wait policy, the module that decides when to wait and when to emit tokens, that does not use the audio.
They use a ``wait-$k$ policy'', historically defined for translating text, which wait for $k$ words of input and then start generating one word of output for each word in the input.
When translating speech, this type of policy instead of counting input words counts chunks of audio.
On standard test sets, which usually have short utterances that are consistently segmented, this may seem to work.
But in the real world, the effects of this type of fixed policy are undesirable.

If the microphone is opened a few seconds before the speaker starts, a wait-$k$ policy forces the LLM to hallucinate output.
If the speaker is slow or hesitates, the system will also produce unnecessary or hallucinated tokens.
If the speaker is fast, the system will fall further and further behind, and start leaving out translations for some phrases.
These are all bad failure modes.

This paper, in contrast, proposes a SpeechLLM for speech-to-text translation that streams adaptively.
It outputs tokens as soon as it has seen sufficient audio to be able to do so.
The translation quality is close to that of the baseline offline system, but the latency is only 1--2 seconds depending on the language, which is much better than existing systems with a fixed policy.
The system in this paper produces \emph{Wait} tokens to indicate that it needs more audio.
In an environment such as a phone, the energy that it costs to produce these extra tokens becomes noticeable.
This paper therefore also proposes an ``early-exit wait policy'', which gets a first chance at deciding whether to emit tokens, or instead to wait.
As this paper will show, this allows a trade-off between latency and energy use.

The paper is structured as follows.
Section \ref{section:architecture} will propose a new architecture for streaming SpeechLLM.
It will also propose an additional wait policy for reducing the energy use on smaller devices.
Key to training a streaming translation system is alignments of the source and target languages, a new method for which will be the topic of Section~\ref{section:alignment}.
Section \ref{section:setup} will give the details of the experimental setup, and propose a new, simpler, metric for the latency.
Finally, Section~\ref{section:experiments} will detail the results.

\subsection{Existing Work on Speech-to-Text Translation}
\label{section:translation:literature}

Traditional speech-to-text translation systems had a \emph{cascaded} architecture, where the best hypothesis from a speech recogniser was piped into a text-to-text translation system \citep{fugen-2007-simultaneous}.
Then, with the advent of encoder-decoder systems for text-to-text machine translation \citep{vaswani-2017-attention}, it became possible to swap the text tokens for speech tokens \citep{gu-2017-learning,seamless-2023,zhang-2024-streamspeech}.
These systems are attractive for a few reasons.
They avoid compounding errors in the speech recognition and the translation.
Also, the non-textual information in the audio could improve the translations \citep{tsiamas-2024-speech}.
Recently, large language models (LLMs) have become popular.
Pretrained LLMs can already perform text-based tasks, which makes it more straightforward, by conditioning them on speech input, to teach them speech recognition \citep{ma-2024-embarrassingly}, or, relevant to this paper, speech-to-text translation \citep{huang-2023-speech, koshkin-2024-transllama,chen-2024-bestow}.

Whether the translation integrates an LLM or not, streaming translation provides an additional challenge.
Since output must be generated before the end of an utterance, the system must decide when to wait for more audio and when there is enough information to output tokens.
This task is performed by a ``wait policy'' \citep{gu-2017-learning}.

A ``wait-$k$ policy'' outputs one word for each input word, which may be useful in a cascaded system \citep{ma-2019-stacl}, where words can be counted.
If, on the other hand, the input is speech, a wait-$k$ policy outputs a token for each chunk of audio \citep{chen-2024-bestow,ouyang-2024-fasst}.
However, in the real world, there is no reason for the speed of the speech to match the wait-$k$ policy's assumption.

Two types of wait policies have only been applied to non-LLM systems.
The first, continuous integrate-and-fire \citep{chang-2022-exploring} is a wait policy that is learned, but with a catch: it is not conditioned on the tokens output so far.
Its task is therefore to decide when tokens should be emitted without knowing the identity of these tokens.
The second, state-of-the-art, wait policy, though it is not learned, is AlignAtt \citep{papi-2023-alignatt}.
AlignAtt requires an encoder-decoder system, so that while streaming, it can examine the cross-attention pattern.
If the attention peak falls within a window of fixed size $\WindowSize$ from the current end of the audio, AlignAtt waits for more audio.
As Section \ref{section:experiments} will show, it degrades the translation quality.

Learned wait policies have been proposed for subtasks of LLM-based speech-to-text translation: ReaLLM \citep{seide-2024-reallm} performs streaming speech recognition and TransLlama \citep{koshkin-2024-transllama} text-to-text translation.
However, streaming speech-to-text translation with a SpeechLLM remains limited to offline processing or relies on a fixed, wait-$k$, policy \citep{chen-2024-bestow,ouyang-2024-fasst}.
This paper will address this gap by introducing a new learned policy for streaming speech-to-text SpeechLLMs.


\section{Architectures for Streaming SpeechLLM}
\label{section:architecture}

A SpeechLLM is an LLM conditioned by speech audio.
The speech audio is usually processed by a pretrained encoder.
First, consider the case where a complete utterance is passed through the encoder and only then the LLM generates output.
There exist two types of architectures to perform the conditioning.
(Appendix~\ref{section:architecture:offline:additional} goes into more detail and gives a mathematical description.)

The first SpeechLLM architecture \citep{ma-2024-embarrassingly} implements the conditioning on speech input by inserting the output of the speech encoder (after passing it through an adapter) as ``speech tokens'' into the LLM's input, before the text tokens.
It will be called a ``concatenated SpeechLLM'', and is one of a class of models sometimes called ``decoder-only''.
The ``intermixed'' architecture that this paper will propose is based on this concatenated architecture but will mix speech tokens and text tokens.

The second SpeechLLM architecture will be called ``cross-attention SpeechLLM'' \citep{chen-2024-bestow}.
It is related to an encoder-decoder architecture, in which the decoder has a stack of self-attention layers and cross-attention layers, where the cross-attention attends to the output of the encoder.
In the cross-attention SpeechLLM, the decoder is a pretrained LLM, so a conditioning network with cross-attention layers are prepended to the LLM.

\subsection{Streaming SpeechLLM: the Intermixed Model}
\label{section:streaming_speechllm}

\begin{figure}
    \begin{subfigure}{0.44\columnwidth}
        \centering
        \includegraphics{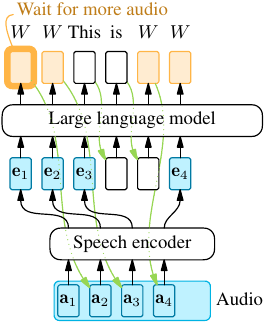}
        \caption{%
            The ``intermixed'' architecture for speech translation.
            The output tokens are intermixed with wait tokens; the input tokens are intermixed with speech tokens.
            When the LLM outputs a text token, that token is passed in at the next step.
            When the LLM outputs a \emph{wait} token $\Wait$, a speech token is passed in at the next step.
            \label{figure:architecture:intermixed}
        }
    \end{subfigure}
    \hspace*{\stretch{1}}
    \begin{subfigure}{0.52\columnwidth}
        \centering
        \includegraphics{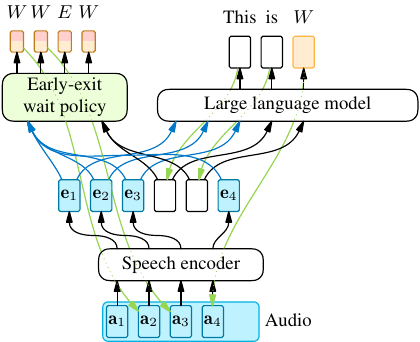}
        \caption{
            The intermixed system from Figure \ref{figure:architecture:intermixed} with an ``early-exit wait policy''.
            When the wait policy decides to wait, the LLM is not evaluated.
            When the wait policy decides to emit, the LLM emits zero or more tokens followed by a wait token.
            This reduces computation, since the wait policy is faster to evaluate than the LLM.
            \label{figure:architecture:intermixed:early-exit}
        }
    \end{subfigure}
    \caption{
        The two architectures proposed in this paper.
        \label{figure:architecture}
    }
\end{figure}

A translation system is streaming if it produces output before the utterance has finished.
For a SpeechLLM, the LLM needs to output tokens conditioned at first on only part of the audio.
As Section~\ref{section:translation:literature} has discussed, existing methods to make SpeechLLM speech-to-text translation streaming use a separate module, which is not learned, as a wait policy.

Instead, this paper proposes a new architecture, the ``intermixed'' SpeechLLM, which integrates a learned wait policy.
The architecture is illustrated in Figure \ref{figure:architecture:intermixed}.
The output tokens, at the top of the figure, intermix normal text tokens with wait tokens $\Wait$.
These are related to blank tokens for the Transducer \citep{graves-2012-sequence} or ReaLLM \citep{seide-2024-reallm} for speech recognition, or wait tokens for text-to-text translation \citep{koshkin-2024-transllama}
In this work, the model outputs a wait token to request another chunk of audio before it can output the next text token.
In the figure, when the LLM outputs a wait token, the next chunk of audio $\AudioVector_{\Time}$ is acquired, and the speech encoder processes it to produce a vector $\EncodedAudioVector_{\Time}$, which is input to the LLM as a ``speech token''.
Each speech token, except for $\EncodedAudioVector_1$, matches a wait token in the previous step.
In practice, multiple speech vectors (in this work, 8) per chunk are passed to the LLM, but only the output after the last one is evaluated.
Either way, the input of the LLM contains two different types of tokens just like the output, but they are text and speech tokens.

For a precise mathematical treatment of streaming SpeechLLM models, see Appendix~\ref{section:streaming:additional}.
Here, it is useful to consider two different types of objects: the sequence $\Words$ of words and the sequence $\Steps$ of steps.
In the example, they are
\begin{subequations}
    \begin{align}
        \Words &= ~ \WordOne ~ \WordTwo ~ \WordThree ~ \EOS \label{eq:eg:words} \\
        \Steps &= ~ \Wait ~ \Wait ~ \WordOne ~ \Wait ~ \Wait ~ \WordTwo ~ \WordThree ~ \EOS \label{eq:eg:steps}%
    \end{align}%
For one word sequence $\Words$, there are exponentially many step sequences $\StepsFor(\Words)$, and the probability of each word sequence in theory is the sum over all matching step sequences.
However, in this work, inference will aim to find the single best step sequence.

Training the intermixed model also uses a single step sequence, by maximising its log-likelihood, or, equivalently, minimising the cross-entropy.
If the model is analysed as having an LLM outputting text tokens and a separate module for the wait policy, then the training criterion is equivalent to multi-task training of the two modules with the same weight, as Appendix \ref{section:training:additional} shows.
Where the $\Wait$ tokens are placed amongst the text token is not fixed; it is decided from the alignment of the speech and the target-language text, which Section~\ref{section:alignment} will detail.

The input $\LLMInput$ to the LLM is also determined by the alignment.
For example, writing the embedding of word $\Word$ as $\Embedding(\Word)$,
\begin{align}
    \LLMInput &= ~ \EncodedAudioVector_1 ~ \EncodedAudioVector_2 ~ \EncodedAudioVector_3 ~ \Embedding(\WordOne) ~ \EncodedAudioVector_4 ~ \EncodedAudioVector_5 ~ \Embedding(\WordTwo) ~ \Embedding(\WordThree). \label{eq:eg:llm_input}%
\end{align}%
\end{subequations}%

Streaming translation models provide a trade-off between latency and translation quality.
To make the intermixed model more or less likely to wait for more audio, this work proposes a ``wait penalty''~$\WaitPenalty$: a value that is subtracted from the log-weight for $\Wait$.
The reason for formulating this as a penalty is that search usually works best when producing wait tokens is mildly  discouraged.
Otherwise, the model will tend to prefer the shorter sequence, i.e.\ only wait tokens, over the longer sequence, i.e.\ text tokens intermixed with wait tokens.

\subsection{Reducing Energy Use on Device: Intermixed with Early-Exit Policy}
\label{section:early-exit}

If the intermixed system is deployed on device, the wait tokens create a potential problem: the increase in the number of LLM calls increases the energy consumption of the system.
(Though the latency is not affected as long as the LLM inference is quicker than the time needed to receive a new audio chunk, typically 640\,ms.)
For this specific scenario, an ``early-exit wait policy'' is proposed.
The resulting architecture is illustrated in Figure \ref{figure:architecture:intermixed:early-exit}.
In the implementation in this paper, the early-exit wait policy is an additional head from the output of the first few layers of the LLM.
The wait policy is faster to run but less sophisticated than the LLM itself, and it has only two outputs: \emph{wait} $\Wait$, which indicates more audio data is needed, or \emph{emit} $\Emit$ to pass control to the LLM.

In inference, the early-exit wait policy is used in a conservative manner.
Appendix~\ref{section:early-exit:additional} goes into more detail.
If it outputs $\Wait$, the LLM is not evaluated and the system waits for the next chunk of audio.
If instead it outputs $\Emit$, the LLM is evaluated repeatedly until it outputs a wait token.
Crucially, the LLM can emit a wait token straight away, in which case nothing is lost except for energy; so the translation quality is not affected.
The early-exit wait policy is parameterised with a separate wait penalty $\WaitPolicyWaitPenalty$, to trade off latency and energy efficiency, but not translation quality.

\section{Phrase-Level Alignment for Streaming Translation}
\label{section:alignment}

\begin{figure}
    \begin{subfigure}{6.90cm}
        \centering
        \includegraphics{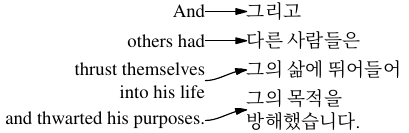}
        \caption{A monotonic alignment.
            \label{figure:alignments:examples:monotonic}
        }
    \end{subfigure}
    \hspace*{\stretch{1}}
    \begin{subfigure}{6.47cm}
        \centering
        \includegraphics{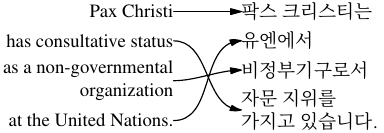}
        \caption{An alignment that involves re-ordering.
            \label{figure:alignments:examples:reordering}
        }
    \end{subfigure}
    \caption{
        Examples of phrase-by-phrase alignments between English and Korean.
        \label{figure:alignments:examples}
    }
\end{figure}

Training data for machine translation does not normally include information about when the system should emit each word.
However, a streaming system needs this.
For generating this training data, the main difficulty is in determining which words in the source-language sequence match which words in the target-language sequence, which is traditionally called an ``alignment''.
Figure~\ref{figure:alignments:examples} illustrates two alignments between English phrases and Korean phrases.
In Figure~\ref{figure:alignments:examples:monotonic}, the alignment is monotonic.
\begin{CJK}{UTF8}{mj} 
By contrast, Figure~\ref{figure:alignments:examples:reordering} illustrates a non-monotonic phrase alignment.
In translation, the Korean ``유엔'' (``UN''/``United Nations'') cannot be generated until English ``United Nations'' is heard, at the end of the utterance.
\end{CJK}

Previous speech-to-text translation studies \citep{koshkin2024transllama, hu-2025-word} have aligned a single word with a single word using methods based on word embedding similarity \citep{sabet2020simalign, dou2021word}.
For languages as different as English and Korean, however, concepts that are one word in one language are not necessarily a single word in the other, making precise word-level alignment challenging.
As Section~\ref{section:experiments} will show, training with word-level alignments produced following \citet{sabet2020simalign} leads to inferior translation quality.
Therefore, this paper proposes to align the source-language and target-language text phrase by phrase, like in Figure~\ref{figure:alignments:examples}.
Here, a phrase is defined as the minimal contiguous span of words that preserves translational equivalence.

The first ingredient required is alignments between the source transcript and the target translation, which this paper proposes to generate with an LLM with few-shot prompting.
For each aligned phrase pair, all words in the target-language phrase are aligned to the final word of the corresponding source-language phrase.
For around 10\,\% of sentences, the LLM mentions a phrase in either language that does not exist, or it omits some text.
This paper proposes a simple heuristic.
Invalid phrase alignments are ignored.
Any target-language text which ends up unaligned is treated conservatively: the affected target phrase is assigned the same alignment as the next valid target phrase.
This heuristic is effective for providing supervision to high-quality streaming translation systems, as Section~\ref{section:experiments} will show.

The other ingredient required is alignments of the audio with words in the source language, which are produced using a speech recogniser.
These can be combined into alignments of audio with phrases in the target language.
The target sequence of wait and text tokens discussed in Section \ref{section:streaming_speechllm} is then computed by requiring that to output a word, the matching audio must have been heard, and that words are produced in order.

\section{Experimental Set-Up}
\label{section:setup}

This section details the model design (Section~\ref{section:models}), the datasets for training, evaluation and test (Section~\ref{section:datasets}), and the metrics (Section~\ref{section:metrics}) used to compare the different streaming and offline SpeechLLM translation systems.

\subsection{Models}
\label{section:models}

The intermixed SpeechLLM is compared to two streaming and two offline baselines based on two different architectures: Bestow \citep{chen-2024-bestow} and a concatenated decoder only SpeechLLM. More precisely, Bestow is used offline, online with a fixed wait-$k$ wait policy and online with a dynamic strategy following AlignAtt \citep{papi-2023-alignatt}. The offline concatenated decoder serves as a high latency and high translation quality upper bar. All models share the same speech encoder and LLM.

\begin{description}

\item[Speech encoder.] Features are extracted using a Conformer model made of 22 layers for a total of 300M of trainable parameters. Prior to the Conformer, a two-layered 2D CNN is employed to downsample the 80 Mel Filterbanks from a frequency of 100 frames per seconds to 25 frames per second. The obtained features are then downsampled once again after the Conformer encoder to 12.5 frames per second using a simple concatenation of consecutive frames. This architecture is first pre-trained for 300,000 steps using self-supervised learning following the BEST-RQ approach \citep{whetten-2024-open} and dynamic chunk training (DCT) with 4 hours of speech per batch \citep{li-2023-dynamic} on the full Loquacious dataset \citep{parcollet-2025-loquacious}. Then, it is fine-tuned on the translation task once plugged to the rest of the system.
During DCT pre-training the chunk size in ms is drawn uniformly from $\{320,640,1280\}$, and during inference, it is fixed to 640\,ms.
A special ``End-of-Audio'' embedding is added to the speech embedding sequence to let the LLM know the end of the audio stream (see Appendix \ref{section:eos} for further details). 

\item[The LLM] is an in-house LLM with 3B parameters. Its parameters are frozen and LoRA of rank 8 is applied to all dense layers during fine-tuning on the translation task. The languages considered for translation in this article are part of the training data of the LLM. 

\item[Generation of alignments]
We generate speech-to-target-language translation alignments using a cascaded procedure.
First, we use the NeMo forced-alignment tool \citep{kuchaiev2019nemo} to obtain word-level timestamps for the source-language transcript.
Next, we prompt Qwen3-14B \citep{yang-2025-qwen3} with five-shot examples to generate phrase-level alignments between the source-language transcript and the target-language translation.

Following \citet{koshkin2024transllama}, we also consider SimAlign \citep{sabet2020simalign}, which uses word embeddings to generate word-level alignments, as a baseline for comparison.
We generate the word-level alignments using the open-source script by \citet{sabet2020simalign}.

\item[The intermixed model] SpeechLLMs integrate the speech encoder and LLM by upscaling speech embeddings to match the LLM's input dimensions via a two-layer dense network, often referred to as modality adapter in the literature, with GeLU activations.

\item[Bestow architecture.] Bestow utilizes a conditioning network that incorporates cross-attention between speech encoder outputs and text embeddings. This network introduces two new Transformer decoder layers at the beginning of the LLM, enabling the text embeddings to be conditioned on the speech input before being processed by the LLM. The network architecture consists of two repetitions of the following sequence: RMS normalization, self-attention, cross-attention, and a feed-forward network, with pre-normalization residual connections at each step. Positional encoding is implemented using RoPE \citep{su-2024-roformer}. Each layer has 8 attention heads and a dimension size of 1,536. One projection layer is added at each end of the conditioning network to first reduce the embedding size from 3,072 to 1,536, and then to upscale it back to 3,072 to reach the expected LLM input dimension. A last residual connection is added to feed the original text embedding to the corresponding conditioned vector right before the LLM. The network is trained from scratch treating tokens as queries and speech embeddings as keys during the speech-to-text translation task. Causal masking is applied to queries while keys are either unmasked for the offline case, or masked according to the ``wait-$k$'' policy. 

\item[The early exit policy] adds an extra head to the intermixed system. In practice, both the intermixed and concatenated baselines share the exact same architecture as the only difference lies in how the input text and speech are given. However, if the early exit policy is used, an extra linear layer is added right after the output of the LLM layer of interest. Following an ablation study, the second layer of the LLM is used as the early exit. The output are simply project to a bi-dimensional decision indicating a wait or an emission. This is trained from scratch with a multi-task objective alongside the rest of the system during translation.

\item[Wait policies.] Both intermixed systems rely on the strategy defined in Section~\ref{section:architecture}. In practice we investigate various wait penalties for the intermixed system and for the early-exit wait policy.
Wait policies for Bestow are ``wait-$k$'', always waiting for 1280\,ms, or AlignAtt.
Then, the model must output one token either every chunk (640\,ms), which gives lower latency but lower quality, or two chunks (1280ms), which gives higher latency but higher quality. For AlignAtt, we strictly follow the description of \citet{papi-2023-alignatt}.
Briefly, AlignAtt wait policy decides to wait or emit depending on the argmax of the cross-attention score between the current token and the speech embeddings.
If the index is within a defined window starting from the last frame, then the policy waits; otherwise, then it emits.
The main parameter is therefore the window size~$\WindowSize$. We explore $\WindowSize \in \{8,10,12,16,32\}$. As the Bestow conditioning network is made of two cross-attention layers, we average across layers and heads the score to obtain the peak attention index.

\end{description}

\label{compute_resources}
The concatenated baseline concatenates the prompt, the speech embeddings, and the translation tokens.
The speech encoder took 10k GPU-hours on A100 GPUs to train and tune.
All Bestow models are trained with the Adam optimiser for 180,000 steps on the CoLiMu dataset (Section~\ref{section:datasets}) with a batch size of total duration 600 seconds on four H100 GPUs. The concatenated baseline and intermixed systems are trained for 120,000 steps. All models are implemented with SpeechBrain \citep{ravanelli-2024-open-source}.

\subsection{Datasets}
\label{section:datasets}

Two language pairs are considered: English to Korean and English to French.
Speech-to-text translation training is done with a concatenation of three datasets named ``CoLiMu'': LibriSpeech \citep{panayotov-2015-librispeech}, CommonVoice \textit{v14.0} \citep{ardila-2020-common_voice}, and MuST-C \citep{cattoni-2021-must-c}, resulting in roughly 3,700 hours of audio.
We chose to combine these three datasets to not only expand the volume of training material but also to leverage their distinct acoustic characteristics. LibriSpeech consists of audiobook recordings, which provide clean read speech. In contrast, CommonVoice is sourced from crowd contributions, where volunteers read sentences online. While also read speech, it includes a wider variety of accents and much more challenging acoustic conditions, as recordings may be made in diverse environments ranging from smartphones in noisy settings to professional microphones. Finally, MuST-C offers more spontaneous yet still prepared speech, as it is derived from TED Talks. Translated text target translations are obtained following the methodology defined in Section~\ref{section:alignment}.

Evaluation is performed with the Korean and French validation and test sets of Fleurs \citep{conneau-2022-fleurs} as well as an internal lectures dataset in Korean.
An extra set, named ``SilFleurs'', is created to highlight the catastrophic failure of Bestow and AlignAtt. in the presence of pauses and noises in the input. This setting is important as streaming speech translation realistically does not happen often on well segmented and clean utterances.
SilFleurs is obtained by prepending 5 seconds of ``noise-free-sound-0683'' from the Musan dataset \citep{snyder-2015-musan} at -20 dB to the Fleurs test set.

\subsection{Metrics}
\label{section:metrics}

The metrics to understand the performance of the translation systems in this paper are both the translation quality and the latency.
Translation quality is measured by the COMET score \citep{rei-2020-comet} with the \textit{wmt22-comet-da} model. The COMET scores assign a translation quality and accuracy score between 0 and 1 (higher is better) given a reference translation, a reference transcription and the predicted translation. 

To evaluate a streaming system, it is important not just to evaluate the translation quality, but also the latency.
The standard latency metric for streaming speech translation is the ``average lagging'' \citep{ma-2019-stacl}, or its length-adaptive variant \citep{papi-2022-over-generation}.
These metrics have a few flaws, originating in their history as a metric for streaming text-to-text translation (after speech recognition) with a wait-$k$ policy.
They give the same weight to long and short emissions; the point where they are 0 depends on the number of tokens in the utterance; their treatment of tokens emitted after the audio has finished is inconsistent.
Appendix \ref{section:latency:additional} details these flaws.

Instead, this paper proposes to use a much simpler latency metric, the \emph{average logical latency}.
This is the average emission time of all tokens, relative to a system that is unrealistically good, which emits tokens spread out evenly over the duration of the audio.

\section{Experiments}
\label{section:experiments}

All streaming models involve a trade-off between latency and translation quality, which can be adjusted through tunable hyperparameters in the wait policies.
This section presents the optimal results observed for each model as well as explores this trade-off by systematically varying the hyperparameters.
Finally, it compares different text-to-text alignment strategies.

\begin{figure}[!t]
    \begin{subfigure}{0.47\textwidth}
        \centering
        \includegraphics{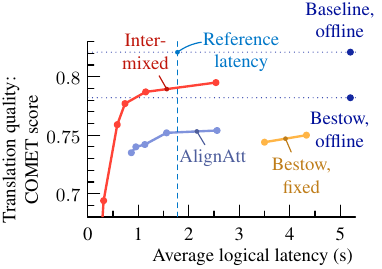}
        \caption{Fleurs: English$\,\rightarrow\,$French.%
        \label{figure:results:en_fr:latency_quality}}
    \end{subfigure}
    \hspace*{\stretch{1}}
    \begin{subfigure}{0.47\textwidth}
        \centering
        \includegraphics{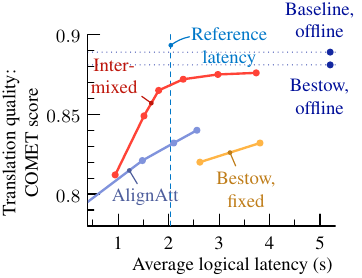}
        \caption{Fleurs: English$\,\rightarrow\,$Korean.%
        \label{figure:results:en_ko:latency_quality}}
    \end{subfigure}
    \caption{Latency vs quality on Fleurs.
        The reference latency is the average logical latency on the alignments (Section \ref{section:alignment}) generated on the test set.
        The curves are generated by varying the wait penaltyfor the intermixed model, window size for Bestow with AlignAtt, and step duration for Bestow with ``wait-$k$'' models.
        Exact parameter values for each point are given in Appendix~\ref{section:experiments:parameters:additional}.
        The intermixed speechLLM offers the best trade-off between latency and quality.
        \label{figure:results:latency_quality}}
\end{figure}

\begin{figure}[!t]
    \begin{subfigure}{0.47\textwidth}
        \centering
        \includegraphics{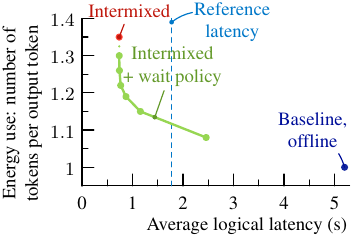}
        \caption{Fleurs: English$\,\rightarrow\,$French.%
        \label{figure:results:en_fr:latency_tokens}}
    \end{subfigure}
    \hspace*{\stretch{1}}
    \begin{subfigure}{0.47\textwidth}
        \centering
        \includegraphics{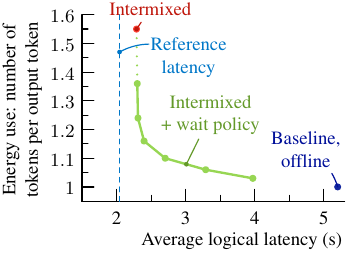}
        \caption{Fleurs: English$\,\rightarrow\,$Korean.%
        \label{figure:results:en_ko:latency_tokens}}
    \end{subfigure}
    \caption{Latency vs energy use on Fleurs.
        The wait penalty is 1 for the LLM, while it varies for the early-exit wait policy.
        The intermixed speechLLM can be tuned to fit the desired energy target.
        \label{figure:results:latency_tokens}}
\end{figure}

\subsection{Results: Latency and Translation Quality}
\label{section:main_results}

Figure \ref{figure:results:latency_quality} details the trade-off between translation quality (COMET) and average logical latency (seconds) across all models.
An ideal SpeechLLM would be situated in the top-left corner, showing high translation quality and low latency.
However, there tends to be a trade-off where reducing the latency negatively affects translation quality.
The intermixed SpeechLLM is the closest to the top-left corner for both English to French (Figure~\ref{figure:results:en_fr:latency_quality}) and English to Korean (Figure~\ref{figure:results:en_ko:latency_quality}), since it has the best latency-quality trade-off of all models.
Compared to the Bestow offline baseline \citep{chen-2024-bestow}, the intermixed model either achieves better (English to French) or comparable (English to Korean) COMET scores at a latency of 1--2 seconds.
Then, relative to the quickest Bestow with a fixed ``wait-$k$'' policy, the latency of the intermixed model, is on average 2.3 times lower while its translation quality is 19.4\% higher.
Since the Bestow architecture uses cross-attention, it is possible to apply AlignAtt \citep{papi-2023-alignatt} to it as a baseline.
This improves the latency significantly and the quality marginally, but it remains inferior to the intermixed approach as seen by its lower position in the graph.
Both Bestow with AlignAtt and the intermixed model obtain the best quality when operating at a latency around the reference latency, demonstrating that they have successfully learned a valid alignment between speech and text for considered language pairs.
These findings extend to English to German speech translation as shown in Appendix \ref{section:experiments:additional_translation}.

However, the intermixed SpeechLLM calls the LLM many more times since it produces extra tokens, the wait tokens.
For energy efficiency on device, the number of extra calls should be reduced.
Adding an early-exit wait policy to the intermixed model allows energy consumption to be tuned.
This does not compromise quality, as the underlying LLM has the ability to override a policy's ``emit'' decision.
Adjusting the wait penalty trades LLM calls for much cheaper and faster wait tokens generated by the early-exit wait policy.
This new trade-off is shown in Figure \ref{figure:results:en_fr:latency_tokens} and Figure \ref{figure:results:en_ko:latency_tokens}.
For all intermixed data points in these curves, the COMET score is the same.
For English to Korean translation, the baseline intermixed system operates at a 2.3\,s latency utilizing 1.55 tokens per output token, i.e.\ the average number of calls to the LLM including wait per non-wait generated token.
Modifying the wait penalty of the early-exit wait policy to 2 and 1 reduces energy consumption to 1.24 and 1.16 tokens per output token, respectively, while holding latencies remarkably stable at 2.3\,s and 2.4\,s.
Reducing the energy consumption further ends up pushing the model closer to the concatenated baseline latency.

\begin{figure}[!t]
    \begin{subfigure}{7.1cm}
        \centering
        \includegraphics{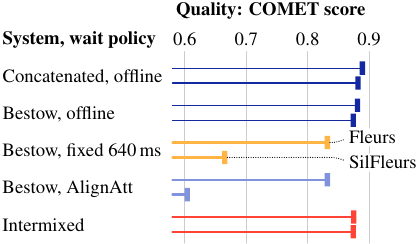}
        \caption{
            Translation quality on Fleurs and SilFleurs.
            The fixed wait policy and AlignAtt fail catastrophically on SilFleurs due to hallucinations from the added silence.
            The intermixed model does not.}
        \label{figure:results:en2ko:quality}
    \end{subfigure}
    \hspace*{\stretch{1}}
    \begin{subfigure}{6.2cm}
        \centering
        \includegraphics{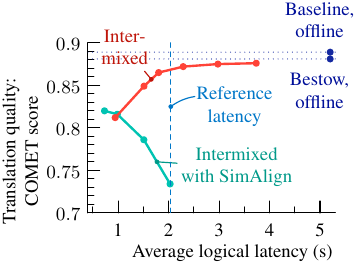}
        \caption{
            Latency vs quality for SimAlign vs this paper.
            SimAlign's word-level alignments are notably less accurate compared to the frame-level alignments proposed in this paper.}
        \label{figure:results:en2ko:simalign}
    \end{subfigure}
    \caption{
        Additional experiments on Fleurs, English$\,\rightarrow\,$Korean.
        \label{figure:results:quality_simalign}
    }
\end{figure}

Finally, Figure \ref{figure:results:en2ko:quality} shows that Bestow with fixed wait policies and AlignAtt exhibit a fundamental vulnerability to acoustic variations that the intermixed systems do not have.
Simply prepending silence to the Fleurs test set (called ``SilFleurs'' in Figure \ref{figure:results:en2ko:quality}) causes Bestow models to fail catastrophically.
For instance, for English to French, COMET scores plummet from 0.820 to 0.509 (wait-$k$ at 640\,ms) and from 0.832 to 0.604 (AlignAtt). Conversely, the intermixed system maintain identical scores despite the extra silence, since streams adaptively.

\subsection{Ablation study: alignment strategy}
\label{section:results:simalign}

Section \ref{section:alignment} has proposed a method for generating the supervision for when the model should emit words.
The key was alignments that are not word by word, but phrase by phrase.
The intuition was the unrelated languages often cannot be matched word by word.
Indeed, we find that median English phrase from the proposed alignment method is 3 words long when matched to Korean phrases, and only 2 words long when matched to French ones.
At the 90th percentile, the difference is even more marked: 9 vs 4 words.

Figure~\ref{figure:results:en2ko:simalign} compares the latency and translation quality of our LLM-generated phrase-level alignments against SimAlign \citep{sabet2020simalign}, a word embedding-based state-of-the-art method for alignments.
For a fixed wait penalty of 1, the intermixed model trained on SimAlign alignments achieve 0.820 COMET and 0.73\,s latency for English-to-Korean translation, which is similar to our phrase-level alignments, which might suggest reasonable performance.
However, Figure~\ref{figure:results:en2ko:simalign} shows that this is not the case.
As with Figure \ref{figure:results:latency_quality}, an ideal system is located at the top-left corner of the graph.
With SimAlign supervision, however, encouraging longer waits degrades the translation quality, which is the opposite of what is expected.
This is also true for English to French, as shown in Appendix \ref{section:experiments:additional}.

We attribute this behaviour to unreliable word-level timing supervision.
SimAlign may align target words to before when the matching source audio has been heard.
The SimAlign-based model is therefore trained to generate prematurely, explaining both the very low latency and the degraded translation quality.
This issue cannot be corrected at inference time by encouraging longer waits, since the underlying supervision does not provide reliable output-timing labels.




\section{Conclusion}
\label{section:conclusion}
This paper has presented a novel approach to build a streaming SpeechLLM system for translating speech in one language into text in another.
The key contribution has been a new SpeechLLM architecture, the ``intermixed'' model, in which the LLM decides whether to wait for more audio or to emit output tokens.
A variant of this architecture the adds an ``early-exit wait policy'' reduces energy use while increasing the latency only slightly.
This paper has proposed a method for aligning the translations with the training audio phrase by phrase, which allows the model to learn when to request more audio.
The new system does not only exhibit much lower latencies than existing fixed and dynamic policies, but also produces translations of higher quality.

\bibliographystyle{icml2025}
\bibliography{literature}

\clearpage
\appendix

\addcontentsline{toc}{section}{Appendix} 
\part{Appendix}
\parttoc

\section{System Architectures}
\label{section:architecture:additional}

\subsection{Offline SpeechLLM}
\label{section:architecture:offline:additional}

\begin{figure}
    \begin{subfigure}{0.48\columnwidth}
        \centering
        \includegraphics{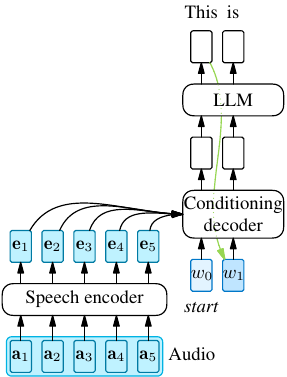}
        \caption{
            The ``cross-attention SpeechLLM'' architecture.
            The way the LLM's output is conditioned on the speech input is by cross-attention.
            To keep the pretrained LLM, cross-attention layers are inserted before the LLM.
            \label{figure:speechllm:cross-attention}
        }
    \end{subfigure}
    \hspace{\stretch{1}}
    \begin{subfigure}{0.48\columnwidth}
        \centering
        \includegraphics{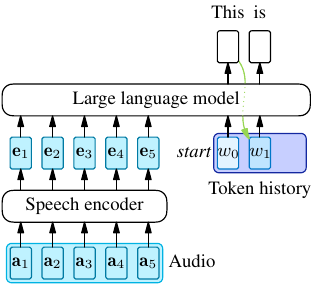}
        \caption{
            The ``concatenated SpeechLLM'' architecture.
            The way the LLM's output is conditioned on the speech input is by feeding the ``speech tokens'' to the decoder as if it were the token history.
            \label{figure:speechllm:concatenated}
        }
    \end{subfigure}
    \caption{
        Conditioning an LLM on speech input in an offline (i.e.\ non-streaming) fashion: the two architectures considered in this work.
        These work for speech recognition or speech translation.
        \label{figure:speechllm}
    }
\end{figure}

Consider two architectures for an offline SpeechLLM, here defined as a non-streaming LLM conditioned on speech input.
The first architecture, here called ``cross-attention SpeechLLM'' \citep{chen-2024-bestow} and shown in Figure~\ref{figure:speechllm:cross-attention}, can be related to the encoder-decoder architecture.
An encoder processes the audio sequence, and a decoder that takes the encoder's output through cross-attention and generates the text output.
There is a difference with an encoder-decoder system, in which all decoder layers have cross-attention.
In the cross-attention SpeechLLM, there is a pretrained LLM, so cross-attention layers are prepended to the LLM.

The second SpeechLLM architecture \citep{ma-2024-embarrassingly} will be called ``concatenated SpeechLLM'', shown in Figure~\ref{figure:speechllm:concatenated},
It is one of a class of models sometimes (confusingly) called ``decoder-only''.
This is because the conditioning on speech input is implemented by inserting the output of the speech encoder as ``speech tokens'' into the LLM's input, before the text tokens.
In this paper, this system will be the basis for a system that mixes speech tokens and text tokens.

Several details of a concatenated system are not visible in Figure~\ref{figure:speechllm}.
First, the speech encoder is pre-trained, in this work with a CTC loss.
There is an adapter between the speech encoder and the LLM, which converts a few outputs of the speech encoder into a token, reducing the frequency to one that is better for the LLM.
The LLM is pretrained, and then fine-tuned, in this work with LoRA, to perform the actual task.
Also Figure~\ref{figure:speechllm}'s single ``\emph{start}'' symbol may in reality be a prompt longer than a single word.

This SpeechLLM architecture can be used for different tasks that require speech input.
The standard task is speech recognition, where the output is text in the same language as the input.
Here, the focus is on using the same architecture for speech-to-text translation, where the language of the output is different from the language that the audio is in.

\subsubsection{Mathematical Description}

Up to this point, the focus has been on the neural architectures of speech-conditioned models.
However, whichever architecture is used, the two SpeechLLM models are mathematically the same.
The mathematical description will be useful to extend to the case of streaming translation.

Call the audio $\Audio={\AudioVector_{1}, \ldots, \AudioVector_{\Length}}$; the encoder $\Encoder(\cdot)$, and the output of the encoder $\EncodedAudio = \Encoder(\Audio)$ or $\EncodedAudioVector_{\Time} = \Encoder(\AudioVector_{\Time})$.
For notational convenience, this assumes that the encoder does not downsample, but it is straightforward to generalise this by introducing different versions of $\Time$.

The output of the model is a token sequence $\Words$, though in the examples, whole-word tokens will be used.
Each token $\Word_{\WordIndex}$ comes from the vocabulary $\Vocabulary$.
Each \emph{complete} token sequence $\Words$ ends with a special end-of-sequence token: $\EOS$.
Where required, it will be made explicit that tokens are represented by embeddings.
The embedding of token $\Word$ will be written $\Embedding(\Word)$.

\begin{figure}
    \centering
    \fbox{\begin{minipage}{0.9\textwidth}
        The overall encoder-decoder model:
        \begin{align*}
            \tilde P(\Words | \Audio) &\DefinedAs
            \prod_{\WordIndex} \Decoder(\Word_{\WordIndex} | \Word_{1}, \ldots, \Word_{\WordIndex-1}, \Encoder(\Audio)),
        \end{align*}
        where these quantities are defined: \\
        \begin{tabular}{ll}
            \hline
            Audio & $\Audio = \{\AudioVector_{1}, \ldots, \AudioVector_{\Length}\}$. \\
            \hline
            Encoded audio & $\EncodedAudio = \Encoder(\Audio) = \{\EncodedAudioVector_{1}, \ldots, \EncodedAudioVector_{\Length}\}$ \\
            \hline
            Conditional probability of token & $\Decoder(\Word_{\WordIndex} | \Word_{1}, \ldots, \Word_{\WordIndex-1}, \EncodedAudio)$ \\
            \hline
        \end{tabular}
    \end{minipage}}
    \caption{\label{figure:eq:encoder-decoder}%
        The model for an offline encoder-decoder system, like either SpeechLLM system from Figure~\ref{figure:speechllm}.
    }
\end{figure}

Figure~\ref{figure:eq:encoder-decoder} shows the mathematical components that a SpeechLLM, uses.
The encoder generates a sequence $\EncodedAudio$ of vectors, and an autoregressive LLM decoder, conditioned on that sequence of vectors, generates tokens one by one.

In the case of a concatenated system, as in Figure~\ref{figure:speechllm:concatenated}, the conditioning is implemented by concatenating the vectors and the embeddings of the token history so far, which can be written as
\begin{align}
    \left[
        \begin{array}{ccccccc}
            \EncodedAudioVector_{1} & \ldots & \EncodedAudioVector_{\Length}
            & \Embedding(\StartToken)
            & \Embedding(\Word_1) & \ldots & \Embedding(\Word_{\WordIndex-1})\\
        \end{array}
    \right]
    .
\end{align}

\subsection{Transducer Model}
\label{section:transducer}

\begin{figure}[t!]
    \centering
    \fbox{\begin{minipage}{0.9\textwidth}
        The overall Transducer model in streaming mode:
        \begin{align*}
            P(\Words| \Audio) &\DefinedAs
                \!\!\!\!
                \sum_{\Steps \in \CompleteStepsFor(\Words, \Length)}
                \!\!\!\!
                \Joiner(\Steps | \Audio)
        \end{align*}

        These quantities are defined: \\
        \begin{tabularx}{\textwidth}{Yl}
            \hline
            Audio up to $\Time$ & $\{\AudioVector_{1}, \ldots, \AudioVector_{\Time}\}$ \\
            \hline
            Encoded audio up to $\Time$ & $\EncodedAudioVector_{1:\Time}
                = \Encoder(\AudioVector_{1:\Time})
                = \{\EncodedAudioVector_{1}, \ldots, \EncodedAudioVector_{\Time}\}$ \\
            \hline
            Predictor (decoder) & $\Decoder(\Word_{1:\WordIndex})
                \in \Reals^{\DecodedSize}$ \\
                \hline
            Joiner, which gives the conditional probability of the next word or $\Wait$
                & \pbox[t]{20em}{
                    $\Joiner(\Step_{\StepIndex} | \Step_{1:\StepIndex-1}, \AudioVector_{1:\Time(\Step_{1:\StepIndex-1})})$\\
                    \hspace*{1em} $= \Joiner(\Step_{\StepIndex} | \Decoder(\Word_{1:\WordIndex(\Step_{1:\StepIndex-1})}), \EncodedAudioVector_{\Time(\Step_{1:\StepIndex-1})} ) $
                } \\
                \hline
            Conditional probability of step sequence &
                $\Joiner(\Steps | \Audio)
                    = \prod_{\StepIndex} \Joiner(\Step_{\StepIndex} | \Step_{1:\StepIndex-1}, \AudioVector_{1:\Time(\Step_{1:\StepIndex-1})})$ \\
            \hline
        \end{tabularx} \\

        The predictor does not depend on the audio, and produces a vector, but not one with logits.
        The predictor also does not depend on the alignment, but just on the word sequence up to the current point.
        
        The joiner therefore also does not depend on the alignment, but just on the word sequence and the audio sequence up to the current point.

        To convert the Transducer to non-streaming, encode all audio $\Audio$ at once, so that the joiner depends on all audio.
    \end{minipage}
    }
    \caption{\label{figure:eq:transducer}%
        The Transducer model for streaming recognition.}
\end{figure}

The Transducer \citep{graves-2012-sequence} is a speech recognition model often used for streaming.
This makes it useful as a comparison.
The general functional form of the Transducer is given in Figure~\ref{figure:eq:transducer}.

There are three important parts of a Transducer: the encoder, the predictor, and the joiner.
The joiner combines the outputs of the other two.
The encoder is a standard encoder but works in a streaming manner.
The predictor takes only the word history, but receives no information about wait symbols $\Wait$.
The joiner depends only on a single output of the encoder, and a single output of the predictor.

The predictor can be implemented in various ways.
In Figure~\ref{figure:eq:transducer}, it is at its most general: $\Decoder(\Word_{1:\WordIndex})$.
The original implementation took the form of a Recursive neural network (RNN).
Define a predictor state $\DecoderState_{\WordIndex}$ as the output of the predictor:
\begin{align}
    \DecoderState_{\WordIndex} &\DefinedAs \Decoder(\Word_{1:\WordIndex}).
\intertext{Now the functional form of the RNN predictor can be made recursive:}
    \Decoder(\Word_{1:\WordIndex}) & \DefinedAs \Decoder(\Word_{\WordIndex}, \DecoderState_{\WordIndex-1})
    .
\end{align}

Similar to the predictor, the joiner in Figure~\ref{figure:eq:transducer} is written in its most general and usual form.
However, there exists a variant called the factorised transducer \citep{xie-2022-factorized} which factorises its emission probability.
A separate module, which here will be called the \emph{wait policy}, will be written $\WaitPolicy$.
It gives a probability $\WaitPolicy(\Emit | \ldots)$ of emitting or the probability $\WaitPolicy(\Wait | \ldots)$ of waiting.
Just like the joiner, it is usually a function of one output of the predictor $\DecoderState$ and one output of the encoder $\EncodedAudioVector$.
The joiner is then factorised as
\begin{align}
    \Joiner(\Step | \DecoderState, \EncodedAudioVector)
    &\DefinedAs
    \begin{cases}
        \WaitPolicy(\Wait |\DecoderState, \EncodedAudioVector)
        & \text{if $\Step = \Wait$;} \\
        \WaitPolicy(\Emit | \DecoderState, \EncodedAudioVector)
        \cdot \NonWaitJoiner(\Step | \DecoderState, \EncodedAudioVector)
        & \text{if $\Step \neq \Wait$.}
    \end{cases}
    \label{eq:transducer:factorised}
\end{align}

\subsection{Streaming SpeechLLM}
\label{section:streaming:additional}

A streaming speech recogniser or speech-to-text translation system outputs tokens as the audio comes in.
However, the wait policy, the module that decides when to wait and when to emit tokens, in current LLM-based speech-to-text translation systems is a ``wait-$\StartDelay$ policy'' \citep[e.g.][]{chen-2024-bestow,ouyang-2024-fasst}.
For translating text, a wait-$\StartDelay$ policy means to wait for $\StartDelay$ words of input and then start generating one word of output for each word in the input.
When translating speech, this type of policy, instead of counting input words, counts chunks of audio.
On standard test sets, which usually have short utterances that are consistently segmented, this may seem to work.
But in the real world, the effects of this type of fixed policy are undesirable.

Instead, this work focuses on a system that learns when to output tokens.
At a high level, to make the model streaming, it must be possible to, just like in the case of the Transducer, explicitly step through time.

\subsubsection{Mathematical Description}

The formalisation that this work chooses is the same as that of the Transducer model (see Section \ref{section:transducer}).
The Transducer augments the output vocabulary $\Vocabulary$ to include a wait symbol $\Wait$, which indicates an explicit move to the next time:
\begin{align}
    \VocabularyWithWait &\DefinedAs \Vocabulary \cup \{ \Wait \}.
\end{align}
(Note that this does not imply that the wait symbol needs to be output by the same component that outputs words, like the Transducer does.)
The model can then output a sequence of steps $\Steps$ that consists of words interspersed with wait symbols~$\Wait$.
For example,
\begin{align}
    \Steps &= ~ \Wait ~ \Wait ~ \WordOne ~ \Wait ~ \Wait ~ \WordTwo ~ \WordThree ~ \EOS \label{eq:eg:steps:additional}
\intertext{Define for convenience $\Words(\Steps)$ that contains just the words in $\Steps$.
    In this example,
}
    \Words(\Steps) &= \WordOne ~ \WordTwo ~ \WordThree ~ \EOS \label{eq:eg:words:additional}
\intertext{Another abstract object that will be useful is the \emph{wait/emit sequence} $\Theta$, which indicates at what times tokens are emitted by interspersing wait symbols $\Wait$ with emit symbols $\Emit$, so that for the example in \eqref{eq:eg:steps:additional}:}
    \WaitEmitSequence &= ~ \Wait ~ \Wait ~ \Emit ~ \Wait ~ \Wait ~ \Emit ~ \Emit \label{eq:eg:wait_emit_sequence}
\intertext{%
Especially for partial hypotheses, it is useful to have shorthands for the following:}
    \WordIndex(\Steps) &\DefinedAs \text{the number of non-$\Wait$ tokens in $\Steps$}; \\
    \Time(\Steps) &\DefinedAs \text{the number of $\Wait$ symbols in $\Steps$}.
\end{align}
In this example $\WordIndex(\Steps) = 4$ and $\Time = 3$, since there are three text tokens and three wait symbols $\Wait$.

In this notation, a non-streaming (or offline) system would always generate all output after having heard the complete audio:
\begin{align}
    \Steps &= ~ \Wait ~ \Wait ~ \Wait ~ \Wait ~ \Wait ~ \WordOne ~ \WordTwo ~ \WordThree ~ \EOS \label{eq:eg:steps:additional:offline}
\end{align}

\begin{figure}
    \centering
    \fbox{\begin{minipage}{0.9\textwidth}
        The overall streaming encoder-decoder model:
        \begin{align*}
            \hat{P}(\Words| \Audio) &\DefinedAs
            \!\!\!\!
            \sum_{\Steps \in \CompleteStepsFor(\Words, \Length)}
            \!\!\!\!
             \Decoder(\Steps | \Audio)
        \end{align*}
        where these quantities are defined: \\
        \begin{tabularx}{\textwidth}{Yl}
            \hline
            Audio & $\Audio = \{\AudioVector_{1}, \ldots, \AudioVector_{\Length}\}$ \\
            Encoded audio at $\Time$ & $\EncodedAudioVector_{\Time}
                = \Encoder(\AudioVector_{1:\Time})$ \\
            Conditional probability of~text token or wait token
            & \pbox[t]{20em}{$
                \Decoder( \Step_{\StepIndex} | \Step_{1:\StepIndex-1}, \AudioVector_{1:\Time(\Step_{1:\StepIndex-1})})
            $ \\
            \hspace*{1em} $
                = \Decoder( \Step_{\StepIndex} | \Step_{1:\StepIndex-1}, \EncodedAudioVector_{1:\Time(\Step_{1:\StepIndex-1})})
            $} \\
            Conditional probability for sequence of steps & $\Decoder(\Steps | \Audio) =
                \prod_{\StepIndex} \Decoder( \Step_{\StepIndex} | \Step_{1:\StepIndex-1}, \AudioVector_{1:\Time(\Step_{1:\StepIndex-1})})
                $ \\
            \hline
        \end{tabularx}\\

    \end{minipage}}
    \caption{The model for a streaming SpeechLLM system.
        \label{figure:eq:encoder-decoder:streaming}
    }
\end{figure}

Figure~\ref{figure:eq:encoder-decoder:streaming} shows the functional form of the streaming encoder-decoder models in this paper.
The encoder is a standard encoder like in Figure~\ref{figure:eq:encoder-decoder} but works in a streaming manner, taking audio $\AudioVector_{1:\Time}$ up to time $\Time$, and outputs encoded audio $\EncodedAudioVector_{1:\Time}$.

The decoder works similarly to the one in the offline encoder-decoder.
The differences are
\begin{itemize}
    \item At each step only the audio up to time $\Time$ is visible.
        The output of the encoder at $\Time$ is $
            \EncodedAudioVector_{\Time} = \Encoder(\AudioVector_{1:\Time})
        $, i.e.\ it only depends on the audio up to time $\Time$.%
        \footnote{%
            Note that with this, $\Decoder(\Steps | \cdot)$ is not a normalised probability, but this work will ignore this issue, since this is true for streaming systems whether for speech recognition or speech translation \citep{variani-2022-global, van_dalen-2025-globally}.
        }
    \item the decoder outputs $\Step \in \VocabularyWithWait$, including wait symbols, instead of only tokens $\Word$.
\end{itemize}

Since there are multiple different sequences of steps corresponding to a specific token sequence, the model marginalises over them.
Call the set of all sequences of steps corresponding to a specific token sequence $\StepsFor(\Words, \Length)$:
\begin{align}
    \StepsFor(\Words, \Length) &\DefinedAs \{ \Steps: \Words(\Steps) = \Words \},
\end{align}
and the set of all sequences of steps corresponding to a specific word sequence that end in $\EOS$ $\CompleteStepsFor(\Words, \Length)$:
\begin{align}
    \CompleteStepsFor(\Words, \Length) &\DefinedAs \{ \Steps: \Words(\Steps) = \Words, \Step_{\lvert \Steps \rvert} = \EOS \},
\end{align}

\begin{figure}
    \centering
    \includegraphics{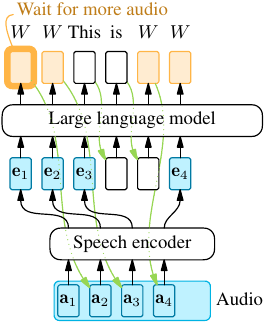}
    \caption{%
        The ``intermixed'' architecture for speech translation, repeated from Figure~\ref{figure:architecture:intermixed}.
        The output tokens are intermixed with wait tokens; the input tokens are intermixed with speech tokens.
        \label{figure:architecture:intermixed:additional}
    }
\end{figure}

It is possible to leave the decision whether to wait or to emit to a separate module, a \empty{wait policy}, which gives a distribution over $\WaitEmitSequence$, and the sequence of text tokens $\Words$.
At each step, the wait policy $\WaitPolicy(\Emit | \ldots)$ gives the probability of emitting at a specific step.
The probability of a step is factorised into the wait policy and a distribution~$\NonWaitDecoder$ over tokens given that a token must be emitted.
\begin{align}
    &\Decoder( \Step_{\StepIndex} | \Step_{1:\StepIndex-1}, \AudioVector_{1:\Time(\Step_{1:\StepIndex-1})})
    \DefinedAs
    \begin{cases}
        \WaitPolicy(\Wait |\Step_{1:\StepIndex-1}, \AudioVector_{1:\Time(\Step_{1:\StepIndex-1})})
        & \text{if $\Step_{\StepIndex} = \Wait$;} \\
        \WaitPolicy(\Emit | \Step_{1:\StepIndex-1}, \AudioVector_{1:\Time(\Step_{1:\StepIndex-1})})
        \cdot \NonWaitDecoder(\Step_{\StepIndex} | \Step_{1:\StepIndex-1}, \AudioVector_{1:\Time(\Step_{1:\StepIndex-1})})
        & \text{if $\Step_{\StepIndex} \neq \Wait$.}
    \end{cases}
    \label{eq:streaming_aed:wait_policy}
\end{align}
The current state-of-the-art for translation with a streaming SpeechLLM uses a wait-$\StartDelay$ policy, forces symbols to be emitted every time step starting from step $\StartDelay$ (originally every input word, for a cascaded system):
\begin{align}
    \WaitPolicy(\Time, \WordIndex) &= \begin{cases}
        1 & \text{if $\Time > \StartDelay + \WordIndex$;} \\
        0 & \text{otherwise,} \\
    \end{cases}
\end{align}
where $\StartDelay$ is the delay until the first symbol is submitted.

Current SpeechLLM models for streaming translation \citep{chen-2024-bestow,ouyang-2024-fasst} use this wait policy.
This is a fixed policy, i.e.\ it does not depend on the audio.
On standard test sets, which usually have short utterances that are consistently segmented, this may seem to work.
But in the real world, the effects of this type of fixed policy are undesirable.

\subsection{The Intermixed Architecture}
\label{section:intermixed}

This paper proposes to implement $\Decoder$ directly as a large language model that is also able to output $\Wait$ tokens, similar to a Transducer \citep{graves-2012-sequence}.
Figure~\ref{figure:architecture:intermixed:additional} illustrates this architecture.
This paper calls this architecture ``intermixed'', since the text tokens and the wait tokens are intermixed.

This architecture is related to ReaLLM \citep{seide-2024-reallm} for speech recognition, or TransLlama \citep{koshkin-2024-transllama} for cascaded speech-to-text translation.
In terms of the input to the LLM, it is similar to FASST \citep{ouyang-2024-fasst}, but that does not output wait tokens.
The LLM can be conditioned on intermixed embeddings $\Embedding_{\StepIndex}$ for text tokens, and encoded audio vectors for $\Wait$ symbols:
\begin{align*}
    \Embedding_{\StepIndex+1} &\DefinedAs
    \begin{cases}
        \Embedding(\Step_{\StepIndex}) & \text{if $\Step_{\StepIndex} \neq \Wait$}; \\
        \EncodedAudioVector_{\Time(\Step_{1:\StepIndex})} & \text{if $\Step_{\StepIndex} = \Wait$}. \\
    \end{cases}
    \intertext{Now the decoder simply depends on a sequence of these embeddings:}
    \Decoder( \Step_{\StepIndex} | \Step_{1:\StepIndex-1}, \AudioVector_{1:\Time(\Step_{1:\StepIndex-1})})
    &=
    \Decoder(\Step_{\StepIndex} | \Embedding_{1:\StepIndex}).
\end{align*}
As shown in Figure~\ref{figure:architecture:intermixed:additional}, the input to the LLM is a mix of text tokens and encoded audio vectors, with the order determined by the wait/emit sequence.
The target output is a mix of wait tokens and text tokens.
What Figure~\ref{figure:architecture:intermixed:additional} does not show is that one chunk contains multiple audio vectors (8 in the experiments in this paper).
For this reason, only tokens at the end of a chunk (e.g.\ ``This is $\Wait$'' in the figure) are included in the loss function.

\subsubsection{Training}
\label{section:training:additional}

The main equation in Figure~\ref{figure:eq:encoder-decoder:streaming} marginalises out over this set of all wait/emit sequences.
However, the loss function in this work involves a single wait/emit sequence found as in Section \ref{section:alignment}.
The intermixed model is trained with the cross-entropy or negated log-likelihood $-\log\Decoder(\Steps | \Audio)$ of a single sequence $\Steps$, like a normal sequence-to-sequence model, since it uses, as per Figure~\ref{figure:eq:encoder-decoder:streaming}, the same LLM for deciding whether to wait and for emitting tokens.
It is not necessary to interleave evaluating the integrated wait policy and the LLM, since both the complete reference wait/emit sequence $\WaitEmitSequence = \{ \WaitEmitStep_1, \ldots \WaitEmitStep_{\Length+\NumWords} \}$ and the complete target token sequence $\Words = \{ \Word_0, \ldots, \Word_{\NumWords} \}$ are available.

If the intermixed model is viewed as integrating a wait policy and a model over text tokens, this form of loss function can be seen as multi-task learning over both models.
The loss function $-\log\Decoder(\Steps | \Audio)$ when \eqref{eq:streaming_aed:wait_policy} substituted into the expression for $\Decoder(\Steps | \Audio)$ in Figure~\ref{figure:eq:encoder-decoder:streaming} becomes
\begin{align}
    -\log\Decoder(\Steps | \Audio)
    &=
        -\sum_{\StepIndex} \log \Decoder( \Step_{\StepIndex} | \Step_{1:\StepIndex-1}, \AudioVector_{1:\Time(\Step_{1:\StepIndex-1})})
    \notag\\&
    =
        -\sum_{\StepIndex}
        \log \WaitPolicy(\WaitEmitStep_{\StepIndex} |\Step_{1:\StepIndex-1}, \AudioVector_{1:\Time(\Step_{1:\StepIndex-1})})
        -
        \sum_{\mathclap{\StepIndex:  \WaitEmitStep_{\StepIndex} \neq \Wait}}
        \log \NonWaitDecoder(\Word_{\StepIndex} | \Step_{1:\StepIndex-1}, \AudioVector_{1:\Time(\Step_{1:\StepIndex-1})})
    .
\end{align}
Here, the left term, the sum of $\WaitPolicy$ terms, is the wait policy on all steps in the wait/emit sequence~$\WaitEmitSequence$.
The right-hand term, the sum of $\NonWaitDecoder$ terms, is the LLM on all tokens in the word sequence~$\Words$.

\subsection{Reducing Energy Use on Device: Intermixed with Early-Exit Policy}
\label{section:early-exit:additional}

\begin{figure}
    \centering
    \includegraphics{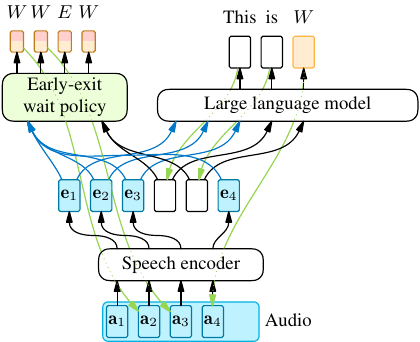}
    \caption{
        The intermixed system from Figure~\ref{figure:architecture:intermixed:additional} with an ``early-exit wait policy'', repeated from Figure~\ref{figure:architecture:intermixed:early-exit}.
        When the wait policy decides to wait, the LLM is not evaluated.
        When the wait policy decides to emit, the LLM emits zero or more tokens and finishes with a wait token.
        This yields a reduction in computation, since the wait policy is faster to evaluate than the LLM.
        \label{figure:architecture:intermixed:early-exit:additional}
    }
\end{figure}

If the intermixed system is deployed on device, the wait tokens create a potential problem: the increase in the number of LLM calls increases the energy consumption of the system, though the latency is not affected.
For this specific scenario, an ``early-exit wait policy'' is proposed.
The resulting architecture is illustrated in Figure~\ref{figure:architecture:intermixed:early-exit:additional}.

Since this wait policy is less sophisticated than the LLM itself, the proposal is to use it in greedy search in a conservative fashion.
The wait policy has only two outputs: \emph{wait} $\Wait$, which indicates more audio data is needed, or \emph{emit} $\Emit$ to pass control to the LLM.
In the latter situation, the LLM can still decide to wait.

The greedy search algorithm for one time step is as follows:
\begin{enumerate}
    \item Evaluate the wait policy.
        If it puts most weight (after subtracting the weight penalty $\WaitPolicyWaitPenalty$) on \emph{wait} $\Wait$, then wait until the next time step.
    \item Otherwise, call the LLM to generate zero (!) or more text tokens.
        If its output puts the highest weight (after subtracting the weight penalty $\WaitPenalty$) on \emph{wait} $\Wait$, then wait until the next time step and run step 1 again.
        Otherwise, output the token with the highest weight.
        Call the LLM again and keep outputting the token with the highest weight every time, until the token is \emph{wait} $\Wait$, in which case go back to step 1.
\end{enumerate}

This search algorithm has these properties:
\begin{itemize}
    \item It never outputs a token before the LLM decides to do so.
        Even if the wait policy puts most weight on \emph{emit} $\Emit$, the LLM can still decide to wait.
    \item It can, however, delay the outputting of tokens, if the wait policy puts most weight on \emph{wait} $\Wait$.
    \item If the wait policy's weight penalty $\WaitPolicyWaitPenalty$ is high, then \emph{emit} $\Emit$ is more likely to be chosen.
        Therefore, the LLM will be called more frequently, and it likely will decide to wait more frequently.
        This does not increase the number of tokens generated, but it does increase the number of LLM calls, and decrease the latency.
        $\WaitPolicyWaitPenalty$~therefore gives a trade-off between energy use and latency.
    \item Assuming the LLM gives the same output if it has been forced to wait for a while without emitting and it would have done if allowed to emit, the translation quality is not affected.
        (This will empirically confirmed by the experiments in Section~\ref{section:experiments}.)
\end{itemize}

The intermixed system with early-exit wait policy is not straightforwardly described in probabilistic terms.
This system will be trained with a multi-task loss, with the cross-entropy loss of the LLM on the intermixed text and wait tokens; and the cross-entropy of the wait policy on the sequence of wait and emit decisions.

The actual architecture of the early-exit wait policy in this work is just the first two layers of the LLM, followed by a linear layer and a softmax into 2 classes: \emph{wait} $\Wait$ and \emph{emit} $\Emit$.

\subsection{End-of-Audio Token}
\label{section:eos}

In principle, streaming systems will continue emitting tokens as long as the input stream generates new features. However, one may want to signal to the model that the speaker has finished speaking, for instance, when the user mutes the microphone in the application.
This could help the model emit the actual ``end-of-sentence'' token, thereby concluding the translation task, or, conversely, prevent it from emitting the token prematurely in the middle of a sentence.
In short, this hints the model to provide a translation of everything it has heard so far.
We have experimented with multiple ``end-of-audio'' tokens, but in practice, it did not make much of a difference in the reported metrics.
This token is implemented as a vector concatenated (in time) to the last speech embedding produced by the speech encoder.
Currently, the intermixed system receives a vector full of zeros. BestOW, on the other hand, receives a random vector (drawn from a normal distribution) generated during training and saved for inference.
Therefore, in both cases, the ``end-of-sentence'' token is distinct from the ``end-of-audio'' token.
During inference, the end of the speech signal triggers a ``flushing'' mechanism.
This means that we first add the ``end-of-audio'' vector to the speech embeddings, and then we force emit until ``end-of-sentence'' is produced. In that case, we ignore any ``wait'' token.
None of these vectors are trained. We did not observe any difference during evaluation with or without it, but further experiments may be necessary.
\section{Ablation Study: Alignment Strategy}
\label{section:experiments:additional}

\begin{figure}[!t]
    \begin{subfigure}{0.47\textwidth}
        \centering
        \includegraphics{figure/latency_quality/latency_quality_en_ko_fleurs_simalign}
        \caption{English$\,\rightarrow\,$French.}
    \end{subfigure}
    \hspace*{\stretch{1}}
    \begin{subfigure}{0.47\textwidth}
        \centering
        \includegraphics{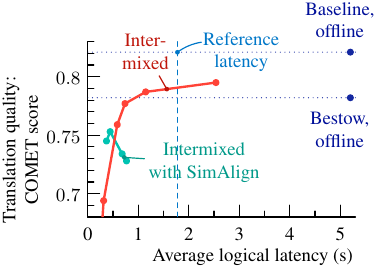}
        \caption{English$\,\rightarrow\,$Korean.}
    \end{subfigure}
    \caption{
        Latency vs quality on Fleurs, comparing the alignment approach in this paper (Section~\ref{section:alignment}) with SimAlign. \label{figure:results:quality_simalign:additional}
    }
\end{figure}

Figure \ref{figure:results:quality_simalign:additional} reports the comparison of our LLM-generated phrase-level alignments against SimAlign \cite{sabet2020simalign} in the same setup as Section~\ref{section:results:simalign} but with English to French Fleurs test set added. The proposed phrase-level alignment offers superior performance to SimAlign and robustness as quality improves as latency increases, while SimAlign translation quality drops significantly.

\section{English to German Speech Translation Results}
\label{section:experiments:additional_translation}

\begin{figure}[!t]
    \centering
    \begin{subfigure}{0.9\textwidth}
        \centering
        \includegraphics{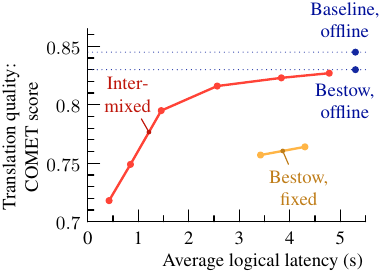}
    \end{subfigure}
    \caption{Latency vs quality on English$\,\rightarrow\,$German. Fleurs.
        The curves are generated by varying the wait penalty ($[-1, 4]$) for the intermixed model and step duration ($[640, 1280]$) for Bestow with ``wait-$k$'' models.
        The intermixed speechLLM offers the best trade-off between latency and quality. \label{figure:results:quality_en2de}}
\end{figure}

Figure \ref{figure:results:quality_en2de} shows the speech translation results obtained by the intermixed model and Bestow with a fixed ``wait-$k$'' policy on the German test set of Fleurs. Models are trained exactly as described in Section \ref{section:setup}. Findings are similar to those of Section \ref{section:experiments} as the intermixed SpeechLLM offers the best trade-off between quality and latency by a significant margin.

\section{Decoding Parameters for Experiments}
\label{section:experiments:parameters:additional}

\begin{figure}[h]
     \centering
     \begin{subfigure}{0.9\textwidth}
         \centering
         \includegraphics{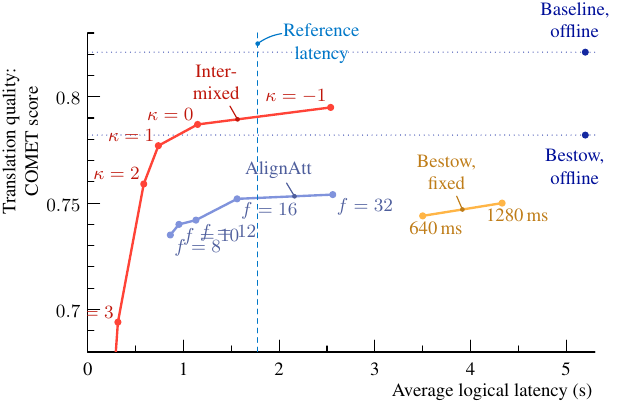}
         \caption{Latency vs quality on Fleurs: English$\,\rightarrow\,$French.}
     \end{subfigure}
     \begin{subfigure}{0.9\textwidth}
         \centering
         \includegraphics{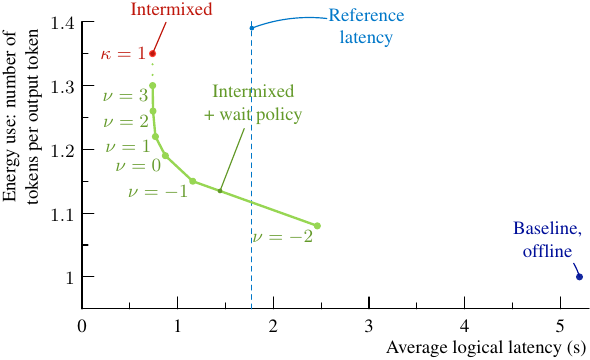}
         \caption{Latency vs energy use on Fleurs: English$\,\rightarrow\,$French.}
     \end{subfigure}
     \caption{Performance with hyperparameters on Fleurs: English$\,\rightarrow\,$French. \label{figure:results:param_details_en2fr}}
 \end{figure}

 \begin{figure}[h]
     \centering
     \begin{subfigure}{0.9\textwidth}
         \centering
         \includegraphics{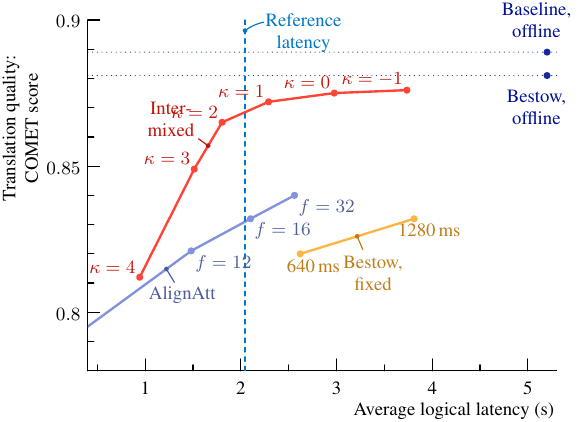}
         \caption{Latency vs quality on Fleurs: English$\,\rightarrow\,$Korean.}
     \end{subfigure}
     \begin{subfigure}{0.9\textwidth}
         \centering
         \includegraphics{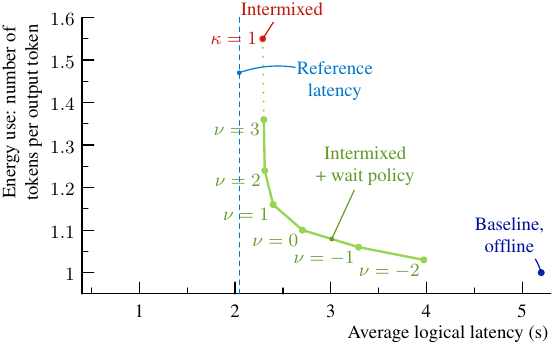}
         \caption{Latency vs energy use on Fleurs: English$\,\rightarrow\,$Korean.}
     \end{subfigure}
     \caption{Performance with hyperparameters on Fleurs: English$\,\rightarrow\,$Korean. \label{figure:results:param_details_en2ko}}
 \end{figure}

 Figure \ref{figure:results:param_details_en2fr} and Figure \ref{figure:results:param_details_en2ko} are detailed views of Figures \ref{figure:results:latency_quality} and \ref{figure:results:latency_tokens} from Section \ref{section:experiments}, annotated with the values of the decoding parameters.
 The results are the same.
\section{Average Logical Latency}
\label{section:latency:additional}

This paper, in Section \ref{section:metrics}, has proposed a new, simple, metric for the latency of a streaming speech-to-text translation system: the average logical latency.
The following will first detail what it is, and Section \ref{section:latency:comparison} will compare it with existing metrics.

The logical latency is measured as the point in the audio where the system is able to produce a token.
It does not take into account the computation time it takes to generate a token.
The latency is relative to a theoretical diagonal line, which imagines tokens are generated at a steady pace from the start to the end of the audio.
The diagonal line represents implausible performance, since most languages have different word orders.

\begin{figure}
    \centering
    \begin{subfigure}{0.9\textwidth}
        \hspace*{2.5cm} 
        \includegraphics{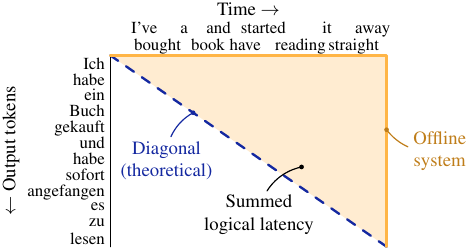}
        \caption{
            The logical latency of an offline (non-streaming) system.
            \label{figure:logical_latency:offline}
        }
    \end{subfigure}
    \\\vspace*{1em}
    \begin{subfigure}{0.9\textwidth}
        \hspace*{2.5cm} 
        \includegraphics{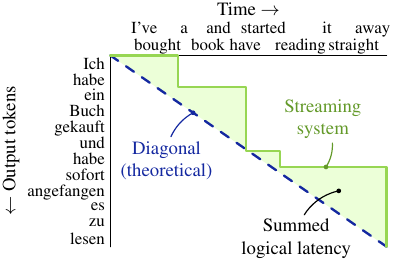}
        \caption{
            The logical latency of a streaming system.
            \label{figure:logical_latency:streaming}
        }
    \end{subfigure}
    \caption{
        The average logical latency illustrated.
        The horizontal axis is time, as the audio of an English speaker comes in; the vertical axis has output words in German.
        The diagonal line is a theoretical baseline.
        The summed logical latency is the area between this baseline and the real system's output timings.
        The average logical latency is the summed logical latency, divided by the number of tokens.
        \label{figure:logical_latency}
    }
\end{figure}

Figure \ref{figure:logical_latency:offline} shows a timeline, just focussing on an offline system.
The audio comes in along the horizontal axis, and symbols are emitted on the vertical axis.
Any vertical line segment indicates an emission.
The offline system in this figure waits until the end of the audio to emit all symbols.
The ``diagonal (theoretical)'' line is a line that is impossible to reach, but is useful as a reference.
It assumes that tokens are emitted at a steady pace from the start until the end of the utterance.
The summed logical latency is the area under the curve between the diagonal line and the actual latency curve.
Dividing this area by the number of tokens yields the average logical latency.

This metric becomes more meaningful in Figure \ref{figure:logical_latency:streaming}, where the system uses actual streaming: vertical segments indicating emissions occur before the end of the utterance.
The summed logical latency again is given by the area under the expectation of the curve in Figure \ref{figure:logical_latency:streaming}, divided by the number of symbols emitted. This latency measurement has the following properties:

\begin{description}
    \item[Logical:]
        it considers time stamps in the audio when the system has enough information that it can emit the next symbol.
        This means that it focuses on the research aspects, not the engineering aspects.
    \item[Average:]
        it assumes that the latency of each token across the dataset counts equally, and therefore averages, with equal weight, the time that each symbol is emitted.
    \item[Relative:]
        as in realistic scenarios there is no supervision for the perfect emission times, the diagonal line in the figure gives an optimistic estimate.
        Whatever baseline one chooses, of course, the measurement changes by a constant additive amount, i.e. this is a mere bias to the statistical metric.
\end{description}

\subsection{Comparison with other metrics: average lagging}
\label{section:latency:comparison}

The main metrics for the speed of streaming translation systems are average lagging \citep{ma-2019-stacl} and its variant length-adaptive average lagging \citep{papi-2022-over-generation}.

The history of these metrics accounts for most of their surprising characteristics.
Adaptive lagging was originally proposed for streaming text-to-text translation with a wait-$k$ policy.
Here, the idea was that as words were coming in (in most cases, from a speech recogniser), the output words would be generated after a fixed lag, and the average lagging would measure this lag, which would be $k$ from the wait-$k$ policy.
The metric was then co-opted for audio input.

\begin{figure}
    \centering
    \includegraphics{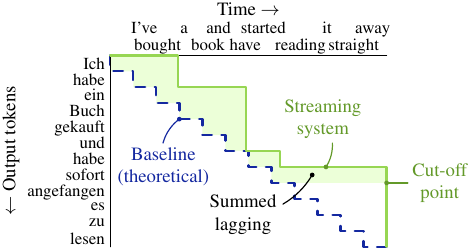}
    \caption{
        The average lagging metric, on the same example as Figure \ref{figure:logical_latency:streaming}.
        \label{figure:average_lagging:streaming}
    }
\end{figure}

Figure~\ref{figure:average_lagging:streaming} illustrates the computation for average lagging.
It is similar to but more complicated than the average logical latency.
The differences are:
\begin{description}
    \item[The baseline] is not a diagonal line but a staircase, which assumes that tokens are emitted half a period sooner than on the diagonal line.
        For very long utterances, the baseline is very close to the diagonal, but for short utterances, it is shifted to earlier in time, which is undesirable.
    \item[The cut-off point] is where average lagging stops counting, which is after the first token emitted at the end of the audio.
        This makes sense only if the wait policy is fixed.
        Otherwise, it has unfortunate effects on the fairness of the calculation, as Section \ref{section:average_lagging:cheating}
    \item[An utterance average] is used (which is not seen in the figure).
        Instead of the behaviour of the system on long utterances having more influence on the final number than on short utterances, average lagging assigns the same weight to each utterance.
        Again, this is of no consequence if the average lagging computed for each utterance is constant, like for wait-$k$ policy.
    \item[Under-generation] and over-generation (with length-adaptive average lagging) are discouraged by the average lagging.
        This is the job of the quality metric, not of the latency metric.
        If the system generates a perfectly clear translation with fewer tokens than the reference, it should not be disadvantaged.
\end{description}

\subsubsection{Average lagging's cut-off encourages not-really-streaming systems}
\label{section:average_lagging:cheating}

\begin{figure}
    \centering
    \begin{subfigure}{0.9\textwidth}
        \centering
        \includegraphics{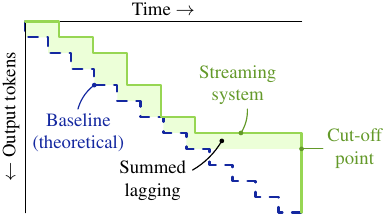}
        \caption{
            An example of emission timings of a streaming system that generates the last few tokens after the end of the audio.
            Average lagging cuts off the calculation after the first tokens after the end of the audio.
            \label{figure:average_lagging_cut_off:original}
        }
    \end{subfigure}
    \\\vspace*{1em}
    \begin{subfigure}{0.9\textwidth}
        \centering
        \includegraphics{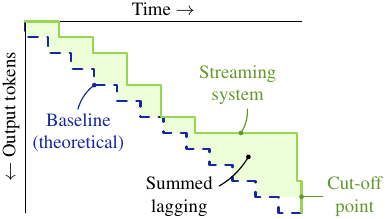}
        \caption{
            An example of emission timings of a streaming system that generates a few of the the last tokens just before the end of the audio.
            Since this does not trigger the cut-off, an extra few tokens are counted in the average lagging calculation.
            The shaded area, the ``summed lagging'' is greater, and the average lagging will be greater.
            This means that the average lagging metric is higher than in Figure \ref{figure:average_lagging_cut_off:original}, though the system actually produces tokens with less lag.
            \label{figure:average_lagging_cut_off:cheating}
        }
    \end{subfigure}
    \caption{
        A case where the cut-off property of average lagging causes unfairness.
        The average lagging calculated for Figure \ref{figure:average_lagging_cut_off:cheating} is higher than that of \ref{figure:average_lagging_cut_off:original}, even though the system generates tokens earlier.
        \label{figure:average_lagging_cut_off}
    }
\end{figure}

The previous section mentioned that the cut-off point for average lagging is the first token emitted at the end of the audio, and that this could lead to unfairness.
Figure \ref{figure:average_lagging_cut_off} illustrates this.
Tokens that are generated just before the end of the audio cause the average lagging metric to increase compared to if the tokens are generated after the end of the audio.
This is a side-effect from not counting tokens after the end of the audio except the first one.
Making systems look better if they give up on streaming and instead wait until the audio has ended is an undesirable property from a latency metric.

\end{document}